\documentclass{article}
\usepackage{ijcai24}
\usepackage[colorlinks=true, linkcolor=blue, citecolor=black, urlcolor=blue]{hyperref}
\usepackage{times}
\usepackage{soul}
\usepackage{url}
\usepackage[utf8]{inputenc}
\usepackage[small]{caption}
\usepackage{graphicx}
\usepackage{amsmath}
\usepackage{amsthm}
\usepackage{booktabs}
\usepackage{algorithm}
\usepackage{algorithmic}
\usepackage[switch]{lineno}

\usepackage{amsthm}
\usepackage{amsmath}
\usepackage{amssymb}   
\usepackage{threeparttable}
\usepackage{algorithm}
\usepackage{algorithmic}
\usepackage{multirow}
\usepackage{graphicx}
\usepackage{arydshln}
\usepackage{booktabs}

\title{Advancing Generalized Transfer Attack with Initialization Derived Bilevel Optimization and Dynamic Sequence Truncation}

\author{
Yaohua Liu$^1$
\and
Jiaxin Gao$^1$\and
Xuan Liu$^{2}$\and
Xianghao Jiao$^{1}$\and
Xin Fan$^1$\And 
Risheng Liu$^{1,3}$\thanks{Corresponding author} 
\\
\affiliations
$^1$School of Software Engineering, Dalian University of Technology, China\\
$^2$CRRC Qingdao Sifang Rolling Stock Research Institute Co., Ltd, China\\
$^3$Pazhou Laboratory (Huangpu), Guangzhou, China.\\
\emails
liuyaohua\_918@163.com,
jiaxinn.gao@outlook.com,
liuxuan\_16@126.com,
jiaoxh0331@outlook.com,
\{xin.fan, rsliu\}@dlut.edu.cn
}

\begin{document}

\maketitle

\begin{abstract}
Transfer attacks generate significant interest for real-world black-box applications by crafting transferable adversarial examples through surrogate models. Whereas, existing works essentially directly optimize the single-level objective w.r.t. the surrogate model, which always leads to poor interpretability of attack mechanism and limited generalization performance over unknown victim models. In this work, we propose the \textbf{B}il\textbf{E}vel \textbf{T}ransfer \textbf{A}ttac\textbf{K} (BETAK) framework by establishing an initialization derived bilevel optimization paradigm, which explicitly reformulates the nested constraint relationship between the Upper-Level (UL) pseudo-victim attacker and the Lower-Level (LL) surrogate attacker. Algorithmically, we introduce the Hyper Gradient Response (HGR) estimation as an effective feedback for the transferability over pseudo-victim attackers, and propose the Dynamic Sequence Truncation (DST) technique to dynamically adjust the back-propagation path for HGR and reduce computational overhead simultaneously. Meanwhile, we conduct detailed algorithmic analysis and provide convergence guarantee to support non-convexity of the LL surrogate attacker. Extensive evaluations demonstrate substantial improvement of BETAK (e.g., $\mathbf{53.41}$\% increase of attack success rates against IncRes-v$2_{ens}$) against different victims and defense methods in targeted and untargeted attack scenarios. The source code is available at \url{https://github.com/callous-youth/BETAK}.
\end{abstract}

\section{Introduction}
Adversarial attack introduces imperceptible yet artificial perturbations into the clean image, potentially leading to incorrect outputs even serious security concerns~\cite{goodfellow2014explaining,xie2017adversarial,jiao2023pearl}. Among these black-box attack techniques, transfer attacks~\cite{dong2018boosting,xie2019improving} generates Adversarial Examples (AE) by attacking a surrogate model, which leads to better attack performance against these victim models with unknown network structures and gradient information. Due to its ability to initiate attacks without requiring direct access to the victim models, transfer attack has garnered extensive attention and research in various real-world applications~\cite{he2023transferable} such as autonomous driving~\cite{deng2020analysis}  and surveillance systems~\cite{alparslan2020adversarial}.

Building upon this foundation, an extensive array of transfer attack methods has been studied, principally including strategies such as input transformation~\cite{wang2021admix,long2022frequency}, momentum-based~\cite{lin2019nesterov,wang2021enhancing,dong2018boosting}, objective-based~\cite{huang2019enhancing} methods, and modification of network structural properties~\cite{wang2023rethinking,guo2020backpropagating}. The above transfer attack methods essentially optimize a single-variable single-level objective w.r.t. surrogate model, thereby overlooking the explicit formulation and evaluation for the attack performance w.r.t. victim models. This deficiency has significantly limited the generalization performance of the AEs across varying victim models. Meanwhile, several studies~\cite{yuan2021meta,fang2022learning} have also focused on designing meta-learning based methods to improve the attack transferability of AEs with model ensembles or data augmentation techniques. Whereas, they always require multiple sampling iterations from the white-box model zoo and pose challenges in integrating with various types of transfer attack methods. Meanwhile, there is a notable lack of interpretability of attack mechanism along with comprehensive theoretical analysis for these methods.

In this work, we establish the \textbf{B}il\textbf{E}vel \textbf{T}ransfer \textbf{A}ttac\textbf{K} (BETAK) framework to address the above limitations. Specifically, by revisiting the optimization process of transfer attack based on its single-level objective of surrogate model, we propose a new initialization derived bilevel optimization paradigm to explicitly reformulate the nested constraint relationship between the Upper-Level (UL) pseudo-victim attacker (w.r.t. perturbation initialization variable) and the Lower-Level (LL) surrogate attacker (w.r.t. perturbation variable). The  LL subproblem corresponds to the same attack objective as the single-level transfer attack, while the UL subproblem essentially reflects the ultimate objective function to achieve better generalization performance over unknown victim models. As for the solution strategy,  we first explicitly calculate the Hyper-Gradient  Response (HGR) as an effective feedback of transferability against one or more pseudo-victim models. Then we further introduce the Dynamic Sequence Truncation (DST) technique to dynamically adjust the backpropagation path for HGR, which also effectively relieves the computation burden. Besides, we provide convergence analysis of the  BETAK framework where the LL subproblem has non-convex propertity due to complex network structures and objectives. Finally, we not only verify the attack performance of BETAK against various victim models and defense methods, but also consider both targeted and untargeted scenarios for thorough evaluation. Comprehensive ablation studies are also conducted to analyze the effectiveness of BETAK and DST technique. The main contributions are summarized as follows:

\begin{itemize}
	\item  Different from directly optimizing a single-variable single-level surrogate objective, we propose the  \textbf{B}il\textbf{E}vel \textbf{T}ransfer \textbf{A}ttac\textbf{K} (BETAK) framework based on the established initialization-derived bilevel optimization paradigm, to analyze and explicitly reformulate the nested constraint relationship between the UL pseudo-victim attacker and the LL surrogate attacker.
	\item Algorithmically, we propose the Hyper-Gradient Response (HGR) estimation as an effective feedback of the transferability against the pseudo-victim models. Meanwhile, we introduce the Dynamic Sequence Truncation (DST) technique to dynamically adjust the back-propagation pathway for HGR computation, simultaneously alleviating the computational burden.
	\item We provide detailed algorithmic analysis of BETAK with convergence guarantees to support non-convex property of surrogate attackers. This demonstrates its compatibility with a broader spectrum of surrogate models, characterized by intricate non-convex network architectures and objective functions.
	\item In-depth experimental analysis reveals that the BETAK framework exhibits substantial improvements against various victim models and defense methods (i.e., $\mathbf{53.41}\%$ increase of attack success rates against the IncRes-v$2_{ens }$ model). The ablation study also validates the effectiveness of DST technique  to enhance performance and relieve computational burden.
\end{itemize}

\section{Methodology}
We first briefly revisit the single-level transfer attack process w.r.t. the surrogate attacker, and then we introduce the initialization derived bilevel optimization paradigm and propose two core techniques of BETAK, including the HGR estimation and DST operation. 
\subsection{Preliminary of Transfer Attack}\label{preliminary}
We first review the single-level problem definition of transfer attacks to evaluate the transferability of AEs. We denote the dataset as $\mathcal{D}=\{\boldsymbol{u}_{i},\boldsymbol{v}_{i}\}_{i=1}^{\mathcal{M}}$, where $\mathcal{D}_{i}=(\boldsymbol{u}_{i},\boldsymbol{v}_{i})$ represents the image and paired label. Then the fundamental objective function of transfer attack could be described as  
\begin{equation}
	\small
	\underset{\boldsymbol{\delta}\in\mathcal{C}}{\operatorname{max}} \text{ }\mathcal{L}_{\mathtt{sur}}(\boldsymbol{\delta};\mathcal{D}_{i},\mathcal{R}),\label{eq:attack}
\end{equation} 
where $\mathcal{L}_{\mathtt{sur}}(\boldsymbol{\delta};\mathcal{D}_{i},\mathcal{R})= \mathcal{L}_{\mathtt{sur}}(\mathcal{R}(\boldsymbol{u}_i+\boldsymbol{\delta}),\boldsymbol{v}_i)\bigr)$, $\mathcal{R}$ denotes the surrogate model. $\boldsymbol{\delta}$ is the perturbation variable optimized based on various attackers subject to $\mathcal{C}=\{\boldsymbol{\delta}|\|\boldsymbol{\delta}\|_{\boldsymbol{\rho}} \leq \boldsymbol{\epsilon}\}$, where $\mathcal{C}$ represents the range of values for the perturbations generated by the attack, and $\rho$ denotes the norm used for attack generation. Typically speaking, we use $\boldsymbol{\delta}_{k}$ to denote the generated perturbation at $k$-th step maximization:
\begin{equation}
		\small
	\boldsymbol{\delta}_{k+1} \leftarrow  \Pi_{\mathcal{C}}\bigl(\boldsymbol{\delta}_k + \alpha \cdot \mathtt{sgn} \nabla_{\boldsymbol{\delta}} \mathcal{L}_{\mathtt{sur}}(\boldsymbol{\delta}_{k};\mathcal{D}_{i},\mathcal{R})\bigr),\label{eq:delta} 
\end{equation}
where $k=0,1,\cdots,K-1$, and $\Pi$ is the projection operation. To this end, various gradient based attackers (single-step~\cite{goodfellow2014explaining} or multi-step~\cite{dong2018boosting}) could be combined to improve attack performance. 

Therefore, existing methods have explored various techniques to facilitate the maximization optimization in Eq.~\eqref{eq:attack} w.r.t. $\boldsymbol{\delta}$ to enhance the transferability of AEs to unknown victim models, denoted as $\mathcal{V}$. In this case, the ultimate objective function of transfer attacks should be written as 
\begin{equation}
		\small
	\underset{\boldsymbol{\delta}_{K}\in\mathcal{C}_{\mathtt{vic}}}{\operatorname{max}} \frac{1}{N} \sum_{n=1}^N\text{ }\mathcal{L}_{\mathtt{vic}_n}\bigl(\boldsymbol{\delta}_{K};\mathcal{D}_{i},\mathcal{V}_{n}\bigr),\label{eq:transfer_attack}
\end{equation}  
where $\mathcal{L}_{\mathtt{vic}_{n}}$ represents the objective of $n$-th victim models, i.e., $\mathcal{V}_{n}$, $n=1, \dots, N$. $\mathcal{C}_{\mathtt{vic}}$ denotes the constraint for the victim model. Typically, $\mathcal{C}_{\mathtt{vic}}$ is set as the same as $\mathcal{C}$ for simplicity.   Note that  this objective reflects the essential goal of transfer attacks, i.e., enhancing the generalization performance of $\boldsymbol{\delta}_{K}$ when transferred to black-box victim models. 

\subsection{Initialization Derived Bilevel Paradigm}

We first introduce $\boldsymbol{\phi}(\boldsymbol{\delta})$ to denote the perturbation generated by certain types of attackers with $\boldsymbol{\delta}$ as the perturbation initialization. When we remove the $\mathtt{sgn}$ operation~\cite{cheng2021fast}, and take single-step gradient ascent as an example, $\boldsymbol{\phi}(\boldsymbol{\delta})$ can be written as 
\begin{equation}
		\small
	\boldsymbol{\phi}(\boldsymbol{\delta})= \Pi_{\mathcal{C}}\bigl(\boldsymbol{\delta} +\alpha \cdot  \nabla_{\boldsymbol{\delta}} \mathcal{L}_{\mathtt{sur}}(\boldsymbol{\delta};\mathcal{D}_{i},\mathcal{R})\bigr). \label{eq:iterative_attack}
\end{equation}
Then we construct the initialization derived bilevel optimization paradigm to capture the nested constrained relationship between Eq.~\eqref{eq:attack} and Eq.~\eqref{eq:transfer_attack},  which can be written as:
\begin{equation}
		\small
	\begin{aligned}
		&\underset{\boldsymbol{\delta}\in\mathcal{C}}{\operatorname{min}} F\bigl(\boldsymbol{\phi}(\boldsymbol{\delta})\bigl),\text{where }F:=-\frac{1}{N} \sum_{n=1}^N\mathcal{L}_{\mathtt{vic}_n}\bigl(\boldsymbol{\phi}^{\star}(\boldsymbol{\delta});\mathcal{D}_{i},\mathcal{V}_{n}\bigr)\\
		s.t.,&\boldsymbol{\phi}^{\star}(\boldsymbol{\delta}):=\underset{\boldsymbol{\phi}\in\mathcal{C}}{\operatorname{argmin}} f\bigl(\boldsymbol{\phi}(\boldsymbol{\delta})\bigl),\text{where } f:=-\mathcal{L}_{\mathtt{sur}}\bigl(\boldsymbol{\phi}(\boldsymbol{\delta});\mathcal{D}_{i},\mathcal{R}\bigr).\label{eq:blo_attack}
	\end{aligned}
\end{equation}
$\boldsymbol{\phi}^{\star}(\boldsymbol{\delta})$ denotes the ‘best' perturbation derived by iteratively solving Eq.~\eqref{eq:iterative_attack}. In the following, the models utilized to construct the UL objective will be referred to as the pseudo-victim attacker. Accordingly, the LL surrogate attacker aims to generate the ‘best' adversarial perturbation to attack the surrogate model, which shares similar motivation with Eq.~\eqref{eq:attack}. While the UL pseudo-victim attacker is optimized to achieve better generalization performance with the perturbation derived from $\boldsymbol{\delta}$. 

Bilevel optimization~\cite{liu2023averaged} is well recognized for modeling and solving various hierarchical applications problems~\cite{liu2024learning,gao2024collaborative}, such as hyperparameter optimization~\cite{franceschi2017forward,liu2021value} and meta learning~\cite{hospedales2020meta,liu2021boml}. The proposed initialization derived bilevel paradigm clearly illustrates the constrained relationship between the UL pseudo-victim attacker and LL surrogate attacker, and offers a new perspective to generate AEs with better generalization capability when transferred to unknown victim models. 

\subsection{Hyper-Gradient Response (HGR) Estimation}

As shown in Eq.~\eqref{eq:blo_attack}, the UL objective continuously optimizes $\boldsymbol{\delta}$, which is defined as the initial state of the LL variable $\boldsymbol{\phi}$. Then we introduce the HGR estimation to update $\boldsymbol{\delta}$ in Eq.~\eqref{eq:blo_attack}. Specifically, we first construct the iterative dynamical system by extending Eq.~\eqref{eq:iterative_attack} as
\begin{equation}
		\small
	\left.\left\{\begin{array}{l}\boldsymbol{\phi}_0(\boldsymbol{\delta})=\boldsymbol{\delta},\\\boldsymbol{\phi}_{k+1}(\boldsymbol{\delta})= \Pi_{\mathcal{C}}\bigl(\boldsymbol{\phi}_{k}(\boldsymbol{\delta}) -\alpha \cdot  \nabla_{\boldsymbol{\phi}} f(\boldsymbol{\phi}_{k}(\boldsymbol{\delta}))\bigr), \end{array}\right.\right.  \label{eq:dynamical_system}
\end{equation}
where $k=1,\cdots,K.$ Therefore, $\boldsymbol{\phi}^{\star}({\boldsymbol{\delta}})$ could be approximated  by
\begin{equation}
		\small
	\boldsymbol{\phi}^{\star}({\boldsymbol{\delta}})\approx\boldsymbol{\phi}_{K}(\boldsymbol{\delta})=\boldsymbol{\phi}_0(\boldsymbol{\delta})-\alpha \cdot\sum_{k=1}^K\nabla_{\boldsymbol{\phi}}f\bigl(\boldsymbol{\phi}_{k-1}(\boldsymbol{\delta})\bigr)\nonumber.\label{eq:dynamical_mapping}
\end{equation}
Following the chain rule, the hyper-gradient w.r.t. $\boldsymbol{\delta}$ can be explicitly calculated with auto differentiation as follows

\begin{equation}
	\small
	\begin{aligned}
		\nabla_{\boldsymbol{\delta}}F\bigl(\boldsymbol{\phi}_K(\boldsymbol{\delta})\bigl)=\bigl(\nabla_{\boldsymbol{\delta}}{\boldsymbol{\phi}_{K}(\boldsymbol{\delta})}\bigr)^\top\nabla_{\boldsymbol{\phi}}F\bigl(\boldsymbol{\phi}_{K}(\boldsymbol{\delta})\bigr).\label{eq:meta_gradient}
	\end{aligned}
\end{equation}

The hyper-gradient contains second-order gradient information as effective feedback of the transferability of $\boldsymbol{\phi}_K({\boldsymbol{\delta}})$. Then we adopt the estimated hyper-gradient to update $\boldsymbol{\delta}$ by

\begin{equation}
		\small
	\begin{aligned}
		\boldsymbol{\delta}= \Pi_{\mathcal{C}}\bigl(\boldsymbol{\delta}-\beta\cdot\mathtt{sgn}\nabla_{\boldsymbol{\delta}}F\bigl(\boldsymbol{\phi}_K(\boldsymbol{\delta})\bigr)\bigr)\nonumber,\label{eq:ul_update}
	\end{aligned}
\end{equation}
where $\beta$ denotes the learning rate.	As the UL and LL objectives converge, $\boldsymbol{\phi}^{\star}(\boldsymbol{\delta})$ is supposed to obtain better generalization performance on the victim models.

In addition, it should be emphasized that we have no access to the gradient information about unknown victim models in the scenario of transfer attacks. To  define the UL victim attacker in Eq.~\eqref{eq:blo_attack}, we assume that one or more arbitrary known models, in addition to $\mathcal{R}$, could serve as pseudo-victim models to imitate the victim models during transfer process. These pseudo-victim models, denoted as $\mathcal{V}_{n}$, inherently satisfy the black-box attack nature for evaluating adversarial transferability of $\boldsymbol{\phi}_{K}$ and update its initialization, i.e., $\boldsymbol{\delta}$, which finally leads to $\boldsymbol{\phi}^{\star}$ with better generalization performance. We also conduct extensive experiments to investigate the influence of pseudo-victim model selection on transferability of AEs.

From another perspective, existing transfer attack methods always obtain $\boldsymbol{\phi}_{K}$ with empirically chosen or randomly initialized $\boldsymbol{\delta}$ in Eq.~\eqref{eq:dynamical_system}, while we estimate the HGR to optimize the initialization used for attacking $\mathcal{R}$, which finally leads to better generalization performance over unknown victim models. We provide more details to discuss the influence of alternative victim models in the ablation part. 
\subsection{Dynamical Sequence Truncation (DST)}

We further introduce the DST operation to facilitate the HGR estimation, which improves the theoretical properties and relieve the hyper-gradient computation burden simultaneously. In Eq.~\eqref{eq:meta_gradient}, the whole historical sequence of $\boldsymbol{\phi}$, i.e., $\{\boldsymbol{\phi}_{k}\}_{k=1}^{K}$, is used to estimate HGR and find better perturbation initialization. Whereas, since the LL subproblem is mostly non-convex due to the complex network structure of surrogate models and varying objectives, there is no guarantee that the obtained $\{\boldsymbol{\phi}_{k}\}_{k=1}^{K}$ could uniformly lead to the convergence of both subproblems. Meanwhile, although $K$ is crucial for proper approximation of $\boldsymbol{\phi}^{\star}(\boldsymbol{\delta})$,  it is always impractical for us to empirically choose ideal fixed $K$ for HGR estimation. 

In this paper, we propose to dynamically truncate the historical sequence in Eq.~\eqref{eq:dynamical_system} according to the change of the UL objective. We first introduce the variable $\tilde{K}$ to replace fixed $K$ for HGR, calculated by
\begin{equation}
	\small
	\tilde{K}:=\arg\max_{1\leq k\leq K}\left\{F(\boldsymbol{\phi}_k(\boldsymbol{\delta}))\right\}.\label{eq:dst}
\end{equation}
Theoretically, $\tilde{K}$ is chosen to minimize the worst case of $\boldsymbol{\phi}_{k}(\boldsymbol{\delta})$ in terms of $F\bigr(\boldsymbol{\boldsymbol{\phi}}_{k}(\boldsymbol{\delta})\bigr)$ so as to ensure that the sequence of solutions for $f$ is consistently optimized with $F$. This idea originates from the non-convex first-order optimization methods~\cite{liu2020investigating,liu2021towards,liu2023augmenting}, and we extend this technique to relax the theoretical properties of BETAK under the LL non-convex scenario. Therefore, $\boldsymbol{\phi}^{\star}(\boldsymbol{\delta})$ is further approximated by $\boldsymbol{\phi}_{\tilde{K}}$, and HGR can be estimated as
\begin{equation}
	\small
	\begin{aligned}
		\nabla_{\boldsymbol{\delta}}F\bigl(\boldsymbol{\phi}_{\tilde{K}}(\boldsymbol{\delta})\bigl)=\bigl(\nabla_{\boldsymbol{\delta}}{\boldsymbol{\phi}_{\tilde{K}}(\boldsymbol{\delta})}\bigr)^\top\nabla_{\boldsymbol{\phi}}F\bigl(\boldsymbol{\phi}_{\tilde{K}}(\boldsymbol{\delta})\bigr).\label{eq:dst_gradient}
	\end{aligned}
\end{equation}
\begin{algorithm}[!t]
	\caption{BETAK Framework}\label{alg:BETAK}
	\begin{algorithmic}[1]
		\REQUIRE UL iteration $T$, LL attack iteration $K$, perturbation $\boldsymbol{\delta}$, LL learning rate $\alpha$ and UL learning rate $\beta$.  
		\STATE Initialize $\boldsymbol{\delta}^0$.
		\FOR {$t=0 \rightarrow T-1$}
		\STATE  $\boldsymbol{\phi}_{0}(\boldsymbol{\delta})=\boldsymbol{\delta}^t$.
		\FOR {$k=0 \rightarrow K-1$}
		\STATE \% \footnotesize{Iterative attack with $\boldsymbol{\phi}_k(\boldsymbol{\delta})$.}
		\STATE $\boldsymbol{\phi}_{k+1}(\boldsymbol{\delta})= \Pi_{\mathcal{C}}\bigl(\boldsymbol{\phi}_{k}(\boldsymbol{\delta}) -\alpha \cdot  \nabla_{\boldsymbol{\phi}}  f\bigl(\boldsymbol{\phi}_{k}(\boldsymbol{\delta})\bigr)\bigr)$.
		\ENDFOR
		\STATE \% \footnotesize{Calculate $\tilde{K}$ with DST technique.}
		\STATE $\tilde{K}=\arg\max_{1\leq k\leq K}\left\{F\bigl(\boldsymbol{\phi}_k(\boldsymbol{\delta})\bigr)\right\}$. \label{outer_loop_1}
		\STATE \% \footnotesize{Calculate the HGR estimation.}
		\STATE $\nabla_{\boldsymbol{\delta}}F\bigl(\boldsymbol{\phi}_{\tilde{K}}(\boldsymbol{\delta})\bigl)=\bigl(\nabla_{\boldsymbol{\delta}}{\boldsymbol{\phi}_{\tilde{K}}(\boldsymbol{\delta})}\bigr)^\top\nabla_{\boldsymbol{\phi}}F\bigl(\boldsymbol{\phi}_{\tilde{K}}(\boldsymbol{\delta})\bigr)$.
		\STATE \% \footnotesize{Update $\boldsymbol{\delta}$ with $\boldsymbol{\phi}_{\tilde{K}}$.}
		\STATE $\boldsymbol{\delta}^{t+1}= \Pi_{\mathcal{C}}\bigl(\boldsymbol{\delta}^t-\beta\cdot\mathtt{sgn}\nabla_{\boldsymbol{\delta}}F\bigl(\boldsymbol{\phi}_{\tilde{K}}(\boldsymbol{\delta})\bigl)\bigr)$.\label{outer_loop_3}
		\ENDFOR
	\end{algorithmic}
\end{algorithm}
It should be noted that since $\tilde{K}$ is always selected from the range of $[1,K]$, the DST technique naturally saves the computation cost for calculating the hyper gradient in Eq.~\eqref{eq:meta_gradient}.  We summarize the BETAK framework with HGR and DST techniques in Alg.~\ref{alg:BETAK}.  By combining HGR estimation with the DST operation, we further establish the convergence guarantee of BETAK framework, which effectively handles the non-convex property of LL surrogate attacker caused by complex network structure and diverse loss functions.

\begin{table*}[htbp]		
	\centering

	\scalebox{1.0}{
		\begin{threeparttable} 
			\small			
			\renewcommand\arraystretch{1.1}
			\setlength{\tabcolsep}{1.1mm}{
				\begin{tabular}{ccccccccccc}
					\toprule[1.2pt] 
					Attacker & Method & Inc-v3$^{*}$ & IncRes-v2$^{*}$ & DenseNet & MobileNet & PNASNet & SENet & Inc-v$3_{ens3}$ & Inc-v$3_{ens4}$ & IncRes-v$2_{ens }$ \\
					\hline \multirow{5}{*}{ PGD } & N/A & 16.34 & 13.38 & 36.86 & 36.12 & 13.46 & 17.14 & 10.24 & 9.46 & 5.52 \\
					\cdashline{2-11}[2pt/5pt]& SGM & 23.68 & 19.82 & 51.66 & 55.44 & 22.12 & 30.34 & 13.78 & 12.38 & 7.90 \\
					\cdashline{2-11}[2pt/5pt]& LinBP & 27.22 & 23.04 & 59.34 & 59.74 & 22.68 & 33.72 & 16.24 & 13.58 & 7.88 \\
					\cdashline{2-11}[2pt/5pt]& Ghost & 17.74 & 13.68 & 42.36 & 41.06 & 13.92 & 19.10 & 11.60 & 10.34 & 6.04 \\
					\cdashline{2-11}[2pt/5pt]& BPA & \underline{35.36}  & \underline{30.12}  &\underline{ 70.70}  & \underline{68.90}  & \underline{32.52}  & \underline{42.02}  & \underline{22.72}  & \underline{19.28}  & \underline{12.40}  \\
					
					\hline	\multicolumn{2}{c}{ BETAK (Ours)}& $\mathbf{53.34}$ & $\mathbf{45.08}$& $\mathbf{74.68}$& $\mathbf{74.54}$ & $\mathbf{42.48}$ & $\mathbf{49.94}$& $\mathbf{30.80}$& $\mathbf{24.90}$ & $\mathbf{17.60}$\\
					\hline
					\multirow{5}{*}{ MI-FGSM } & N/A & 26.20 & 21.50 & 51.50 & 49.68 & 22.92 & 30.12 & 16.22 & 14.58 & 9.00 \\
					\cdashline{2-11}[2pt/5pt] & SGM & 33.78 & 28.84 & 63.06 & 65.84 & 31.90 & 41.54 & 19.56 & 17.48 & 10.98 \\
					\cdashline{2-11}[2pt/5pt] & LinBP & 35.92 & 29.82 & 68.66 & 69.72 & 30.24 & 41.68 & 19.98 & 16.58 & 9.94 \\
					\cdashline{2-11}[2pt/5pt]& Ghost & 29.76 & 23.68 & 57.28 & 56.10 & 25.00 & 34.76 & 17.10 & 14.76 & 9.50 \\
					\cdashline{2-11}[2pt/5pt] & BPA & \underline{ 47.58}  & \underline{ 41.22}  & \underline{ 80.54}  & \underline{ 79.40}  & \underline{ 44.70}  & \underline{ 54.28}  & \underline{ 32.06}  & \underline{ 25.98}  & \underline{ 17.46}  \\
					
					\hline	\multicolumn{2}{c}{ BETAK (Ours)} & $\mathbf{58.24}$ & $\mathbf{47.30}$& $\mathbf{81.82}$& $\mathbf{80.50}$ & $\mathbf{49.26}$ & $\mathbf{55.76}$& $\mathbf{35.82}$& $\mathbf{31.04}$ & $\mathbf{22.68}$ \\ 
					\hline
					\multirow{5}{*}{ VMI-FGSM } & N/A & 42.68 & 36.86 & 68.82 & 66.68 & 40.78 & 46.34 & 27.36 & 24.20 & 17.18 \\
					\cdashline{2-11}[2pt/5pt]& SGM & 50.04 & 44.28 & 77.56 & 79.34 & 48.58 & 56.86 & 32.22 & 27.72 & 19.66 \\
					\cdashline{2-11}[2pt/5pt] & LinBP & 47.70 & 40.40 & 77.44 & 78.76 & 41.48 & 52.10 & 28.58 & 24.06 & 16.60 \\
					\cdashline{2-11}[2pt/5pt]  & Ghost & 47.82 & 41.42 & 75.98 & 73.40 & 44.84 & 52.78 & 30.84 & 27.18 & 19.08 \\
					\cdashline{2-11}[2pt/5pt]  &  BPA & \underline{ 55.00 } & \underline{ 48.72 } & \underline{ 85.44}  & \underline{ 83.64}  & \underline{ 52.02}  & \underline{ 60.88 } & \underline{ 38.76}  & \underline{ 33.70 } & \underline{ 23.78}  \\
					\hline	\multicolumn{2}{c}{ BETAK (Ours)} & $\mathbf{61.18}$ & $\mathbf{52.18}$& $\mathbf{85.54}$& $\mathbf{84.64}$ & $\mathbf{54.18}$ & $\mathbf{61.44}$& $\mathbf{40.84}$& $\mathbf{34.3}$ & $\mathbf{26.16}$ \\ 
					\hline
					\multirow{6}{*}{ ILA } & N/A & 29.10 & 26.08 & 58.02 & 59.10 & 27.60 & 39.16 & 15.12 & 12.30 & 7.86 \\
					\cdashline{2-11}[2pt/5pt] & SGM & 35.64 & 32.34 & 65.20 & 71.22 & 34.20 & 46.72 & 17.10 & 13.86 & 9.08 \\
					\cdashline{2-11}[2pt/5pt]& LinBP & 37.36 & 34.24 & 71.98 & 72.84 & 35.12 & 48.80 & 19.38 & 14.10 & 9.28 \\
					\cdashline{2-11}[2pt/5pt] & Ghost & 30.06 & 26.50 & 60.52 & 61.74 & 28.68 & 40.46 & 14.84 & 12.54 & 7.90 \\
					\cdashline{2-11}[2pt/5pt]& BPA  & \underline{ 47.62}  & \underline{ 43.50}  & \underline{ 81.74}  & \underline{ 80.88}  & \underline{ 47.88}  & \underline{ 60.64 } & \underline{ 27.94}  & \underline{ 20.64}  & \underline{ 14.76}  \\
					\hline	\multicolumn{2}{c}{ BETAK (Ours)} & $\mathbf{57.14}$ & $\mathbf{50.58}$& $\mathbf{83.28 }$& $\mathbf{82.26}$ & $\mathbf{51.88}$ & $\mathbf{62.06}$& $\mathbf{31.48}$& $\mathbf{23.12}$ & $\mathbf{17.04}$ \\
					\hline
					\multirow{5}{*}{ SSA } & N/A & 35.78 & 29.58 & 60.46 & 64.70 & 25.66 & 34.18 & 20.64 & 17.30 & 11.44 \\
					
					\cdashline{2-11}[2pt/5pt] & SGM & 45.22 & 38.98 & 70.22 & 78.44 & 35.30 & 46.06 & 26.28 & 21.64 & 14.50 \\
					\cdashline{2-11}[2pt/5pt] & LinBP & 48.48 & 41.90 & 75.02 & 78.30 & 36.66 & 49.58 & 28.76 & 23.64 & 15.46 \\
					\cdashline{2-11}[2pt/5pt] & Ghost & 36.44 & 28.62 & 61.12 & 66.80 & 24.90 & 33.98 & 20.58 & 16.84 & 10.82 \\
					\cdashline{2-11}[2pt/5pt] & BPA  & \underline{ 51.36}  & \underline{ 44.70}  & \underline{ 76.24}  & \underline{ 79.66}  & \underline{ 39.38}  & \underline{ 50.00}  & \underline{ 32.10}  & \underline{ 26.44} & \underline{ 18.20} \\
					\hline	\multicolumn{2}{c}{ BETAK (Ours)} & $\mathbf{64.38}$ & $\mathbf{53.14}$& $\mathbf{83.24 }$& $\mathbf{82.92}$ & $\mathbf{54.24}$ & $\mathbf{59.92}$& $\mathbf{41.52}$& $\mathbf{36.08}$ & $\mathbf{27.92}$ \\
					\bottomrule[1.2pt] 
				\end{tabular}
			}	
		\end{threeparttable}
	}
		\caption{We compare the ATR (\%) results based on 5 iterative attackers (i.e., PGD, MI-FGSM, VMI-FGSM, ILA and SSA) incorporated with 4 model based methods (i.e., SGM, LinBP, Ghost and BPA). The AEs are generated with ResNet-50 backbones and tested on 2 pseudo-victim (marked with ${*}$) and 7 victim models. The best and second-best outcomes are designated with boldface and underline, respectively.}
	\label{tab:AT_results1}
\end{table*}
\begin{figure*}[htbp]
	\begin{center}
		\renewcommand\arraystretch{0.1}
		\begin{tabular}{c@{\extracolsep{0.1em}}c@{\extracolsep{0.1em}}c@{\extracolsep{0.1em}}}
			&&\\
			&\includegraphics[height=3.25cm,width=8.25cm,trim=10 0 0 0,clip]{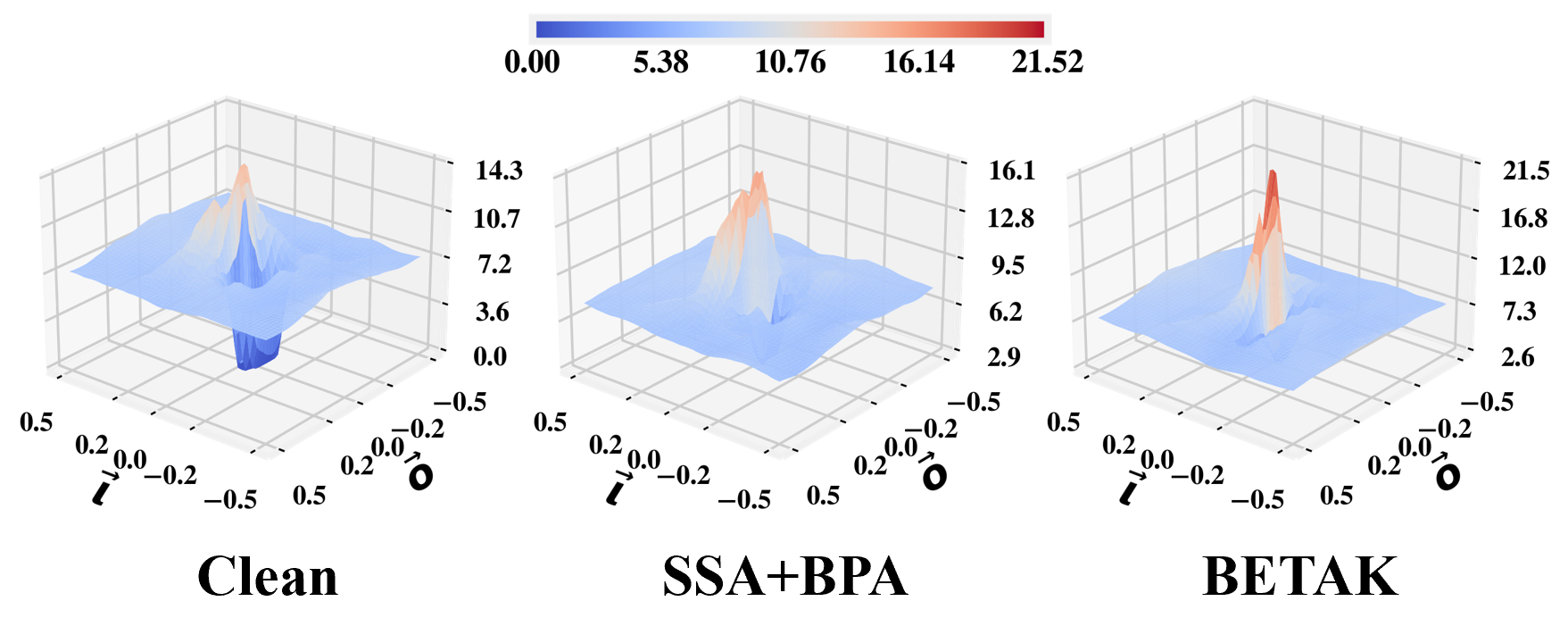} &\includegraphics[height=3.25cm,width=8.25cm,trim= 0 0 0 0,clip]{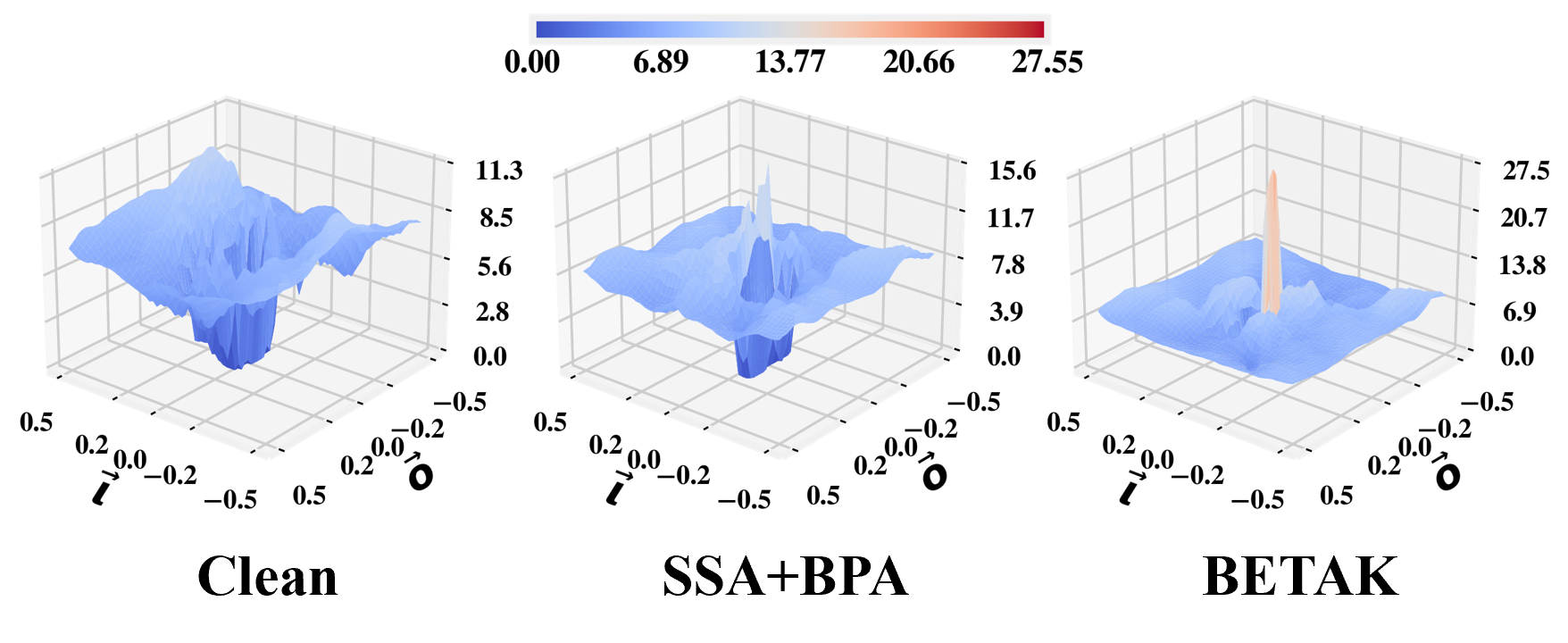}	\\
			&\footnotesize (a) Surrogate Model &\footnotesize (b) Inc-v3 \\
			&\includegraphics[height=3.25cm,width=8.25cm,trim= 10 0 0 0,clip]{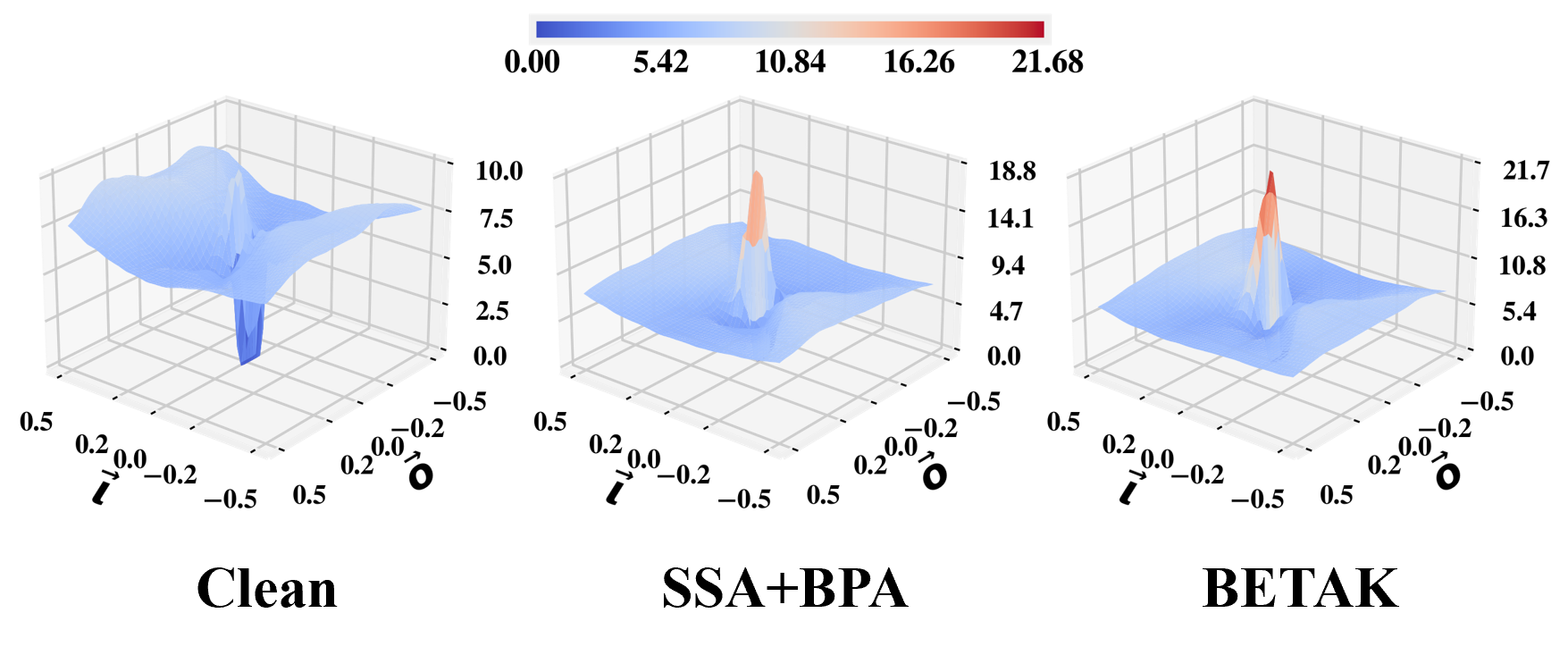}&	\includegraphics[height=3.25cm,width=8.25cm,trim= 0 0 0 0,clip]{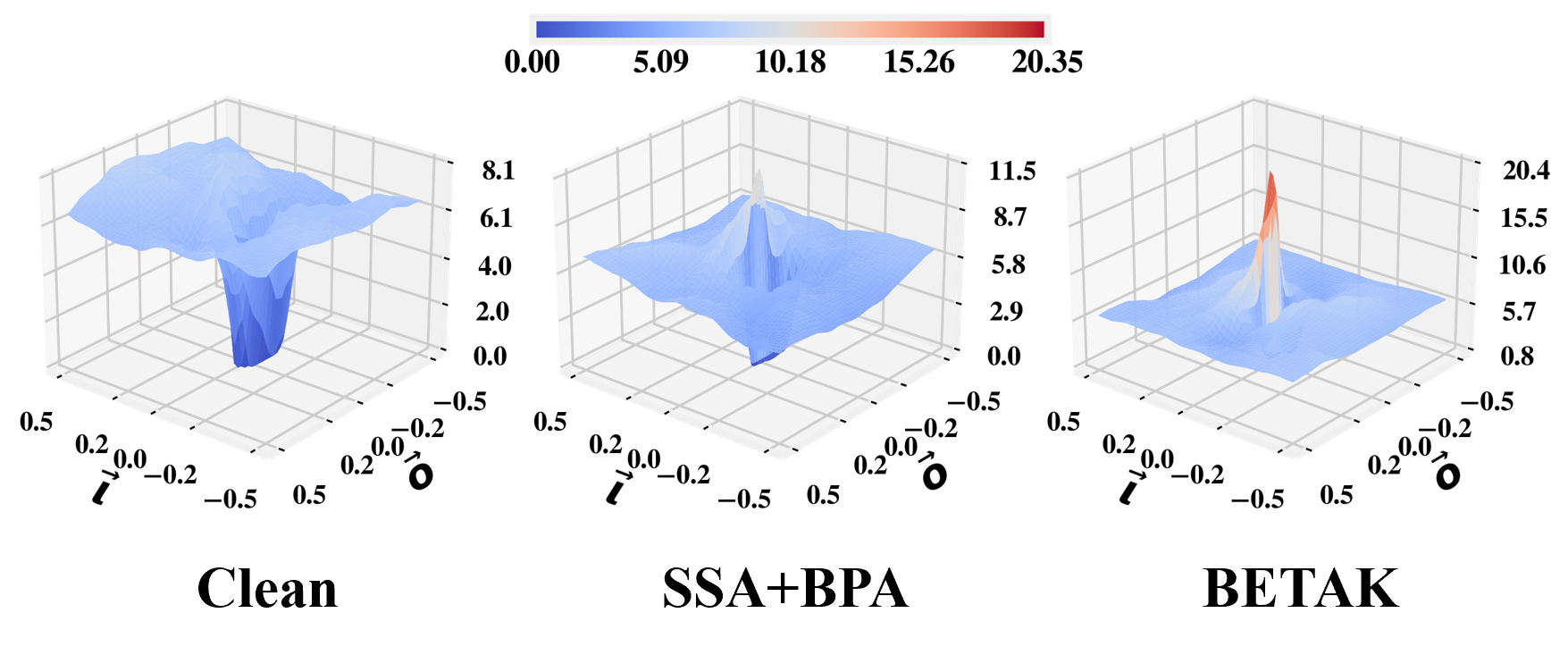}\\
			&\footnotesize (c) DenseNet &\footnotesize (d) MobileNet-v2 \\
		\end{tabular}
	\end{center}
	\caption{The three subfigures in group (a)-(d) illustrate the adversarial loss landscape w.r.t. input variations of the clean sample in Fig.~\ref{fig:fig1} and corresponding AEs perturbed by SSA+BPA and BETAK. The color bars represent a common loss range and color mapping in each group. We mark the maximum and minimum value within the range of $x, y\in[-0.5,0.5]$ w.r.t. $\boldsymbol{\vec{\iota}}$ and $\boldsymbol{\vec{o}}$, respectively. }\label{fig:loss_landscape}
\end{figure*}

\subsection{Algorithmic Analysis of BETAK}

We  delve into the theoretical foundations of BETAK, focusing particularly on the convergence properties in non-convex scenarios. Following previous assumptions~\cite{liu2021towards}, when the UL variables only contain the initialization variable of $\boldsymbol{\phi}_k$, i.e., $\boldsymbol{\delta}$, without including additional parameters, we have that the initialization-based bilevel optimization paradigm, depicted in Eq.~\eqref{eq:blo_attack}, is actually a special case of the explicit gradient-based methods~\cite{liu2023augmenting}. Consequently, with the LL non-convex property, by introducing $\tilde{K}$ to dynamically truncate the historical sequence (i.e., $\{F\bigl(\boldsymbol{\phi}_k(\boldsymbol{\delta})\bigr)\}_{k=1}^K$) obtained from the  gradient based attack iterations, we demonstrate that any limit point of the UL sequence is guaranteed to converge to a stationary point satisfying the first-order optimality conditions of  Eq.~\eqref{eq:blo_attack}. This above convergence analysis ensures that despite the non-convex nature of LL surrogate attackers, BETAK can reliably find solutions that are robust and practically effective to improve the transferability across unknown victim models. When we derive the optimized perturbation initialization, BETAK could flexibly combined various attack methods to perform the surrogate attack process and obtain the optimal perturbation.  In the experimental part, we demonstrate the flexibility of BETAK combined with different basic iterative attackers. 

\section{Experiments}
We evaluate the performance improvement of BETAK compared with representative transfer attack methods against various victim models and defense techniques under both targeted and untargeted attacks.

\subsection{Experimental Setting}

\textbf{Dataset and Models.} We follow the commonly used settings to randomly select 5000 images from 1000 categories of the ImageNet (ISLVRC2012) validation set, all of which could be correctly classified by the victim models. For fair comparison, we choose the commonly used ResNet-50 backbone as the surrogate model. We select 9 victim models with different network structures, including 6 standardly trained models, i.e., Inception (Inc)-v3~\cite{szegedy2016rethinking}, IncRes-v2~\cite{szegedy2017inception}, DenseNet~\cite{huang2017densely}, MobileNet-v2~\cite{sandler2018mobilenetv2}, PNASNet~\cite{liu2018progressive}, SENet~\cite{hu2018squeeze} and 3 robustly trained models, i.e., Inc-v3$_{ens3}$, In-v3$_{ens4}$ and IncRes-v2$_{ens}$~\cite{tramer2017ensemble}. As for the implementation of BETAK, we consider Inception-v3 and IncRes-v2 as the pseudo-victim model since they are proposed earlier. Besides, it can be clearly observed from the ablation results that introducing either one as the pseudo model will significantly improve the attack performance against other victim models.  We adopt the Attack Success Rate (ATR) as the evaluation criterion for the performance of transfer attacks.

\textbf{Baselines and Training Details.} We choose 5 representative attackers including the basic multi-step attacker (i.e., PGD~\cite{kurakin2018adversarial}), 2 momentum based attackers (i.e., MI-FGSM~\cite{dong2018boosting} and VMI-FGSM~\cite{wang2021enhancing}), objective based attacker (i.e., ILA~\cite{huang2019enhancing}) and input transformation based attacker ( i.e., SSA~\cite{long2022frequency}). Besides, we also consider various model-related methods including SGM~\cite{wu2020boosting}, LinBP~\cite{guo2020backpropagating}, Ghost~\cite{li2020learning} and BPA~\cite{wang2023rethinking}. We follow the hyper parameter setting in BPA to implement the above methods. In the untargeted scenario, we set the UL iteration $T=10$, $\boldsymbol{\epsilon}=8/255$ and UL learning rate $\beta=1.6/255$ for all the methods. As for the targeted attacks, we set $T=300$, $\boldsymbol{\epsilon}=16/255$ and $\beta=16/255$. As for the hyper parameter of BETAK, since we omit the $\mathtt{sgn}$ operation, we set the LL learning rate $\alpha=2.0$ and LL iteration $K=10$.

\subsection{Experimental Results}

\textbf{Quantitative Results with Untargeted Attacks.} We first evaluate the attack performance of BETAK against various victim models in comparison with state-of-the-art attack methods. In Tab.~\ref{tab:AT_results1}, we test the ATR with 2 pseudo-victim models and 7 victim models. As it is excepted, BETAK improves the ATR on Inception-v3 and IncRes-v2, since these pseudo-victim models provide the feedback of generalization performance by participating in the HGR estimation w.r.t. $\boldsymbol{\delta}$. More importantly, it is shown that BETAK also significantly improves the attack performance against other 7 victim models.  Therefore, by continuously optimizing the pseudo-victim attacker w.r.t. $\boldsymbol{\delta}$, the generated AEs are supposed to have better generalization performance to unknown victim models. 
\begin{table*}[htbp]
	\centering

	\scalebox{0.8}{
		\begin{threeparttable}   
			\renewcommand{\arraystretch}{1.2} 			
			\setlength{\tabcolsep}{1.1mm}{
				\begin{tabular}{ccccccccccc}
					\toprule[1.2pt] 
					Attacker & Method & Inc-v3$^{*}$ & IncRes-v2$^{*}$ & DenseNet & MobileNet & PNASNet & SENet & Inc-v$3_{ens3}$ & Inc-v$3_{ens4}$ & IncRes-v$2_{ens }$ \\
					\hline & N/A & 0.54 & 0.80 & 4.48 & 2.04 & 1.62 & 2.26 & 0.18 & 0.08 & 0.02 \\
					\cdashline{2-11}[2pt/5pt] 
					& SGM & 2.56 & 3.12 & 15.08 & 8.68 & 5.78 & 9.84 & 0.62 & 0.18 & 0.04 \\
					\cdashline{2-11}[2pt/5pt] 
					PGD & LinBP & 5.30 & 4.84 & 16.08 & 8.48 & 7.26 & 7.94 & 1.50 & 0.54 & 0.28 \\
					\cdashline{2-11}[2pt/5pt] 
					& Ghost & 1.34 & 2.14 & 10.24 & 4.74 & 3.90 & 6.64 & 0.36 & 0.16 & 0.10 \\
					\cdashline{2-11}[2pt/5pt] 
					& BPA & \underline{6.36} & \underline{7.80} & \underline{23.30}& \underline{12.26} &\underline{12.54} & \underline{11.86} & \underline{1.72} & \underline{0.82} & \underline{0.50} \\
					\cline{1-11}
					\multicolumn{2}{c}{ BETAK (Ours)}& $\mathbf{7.66}$ & $\mathbf{9.14}$& $\mathbf{24.90}$& $\mathbf{13.08}$ & $\mathbf{13.78}$ & $\mathbf{12.62}$& $\mathbf{2.08}$& $\mathbf{0.84}$ & $\mathbf{0.70}$ \\
					\hline \multirow{5}{*}{ MI-FGSM } & N/A & 0.16 & 0.26 & 2.06 & 0.90 & 0.42 & 1.22 & 0.00 & 0.02 & 0.02 \\
					\cdashline{2-11}[2pt/5pt] 
					& SGM & 0.74 & 0.76 & 5.84 & 3.24 & 1.66 & 3.70 & 0.00 & 0.02 & 0.00 \\
					\cdashline{2-11}[2pt/5pt] 
					& LinBP & 3.30 & 3.00 & 13.44 & 6.26 & 5.50 & 7.18 & 0.30 & 0.10 & 0.02 \\
					\cdashline{2-11}[2pt/5pt] 
					& Ghost & 0.66 & 0.76 & 5.48 & 2.14 & 1.58 & 3.38 & 0.08 & 0.02 & 0.00 \\
					\cdashline{1-11}[2pt/5pt] 
					& BPA & \underline{4.26}& \underline{5.02} & \underline{18.54} & \underline{8.98} & \underline{8.40} & \underline{10.18} & \underline{0.44} & \underline{0.16} & \underline{0.06} \\
					
					\cline{1-11}
					\multicolumn{2}{c}{ BETAK (Ours)}& $\mathbf{6.62}$ & $\mathbf{6.86}$& $\mathbf{20.06}$& $\mathbf{10.60}$ & $\mathbf{9.52}$ & $\mathbf{10.80}$& $\mathbf{0.76}$& $\mathbf{0.26}$ & $\mathbf{0.08}$ \\
					\bottomrule[1.2pt] 
				\end{tabular}
			}
			\end{threeparttable}
		}
			\caption{We report comparative results of targeted ATR (\%) based on 9 victim models. All the AEs are generated with ResNet-50 backbones. The best and second-best outcomes are designated with boldface and underline, respectively.
		} \label{tab:targeted_attack}
	\end{table*} 
	
	\begin{table}[htbp]
		\centering

		\scalebox{0.8}{
			\begin{threeparttable}   
				
				\renewcommand{\arraystretch}{1.2} 			
				\setlength{\tabcolsep}{1.1mm}{
					\begin{tabular}{ccccccc}
						\toprule[1.2pt] 
						Attacker & Method & HGD & R\&P & NIPS-r3 & JPEG & \multicolumn{1}{c}{ RS } \\
						\hline \multirow{5}{*}{ PGD } & N/A & 17.40& 5.58 & 14.58 & 11.42 & 8.00  \\
						\cdashline{2-7}[2pt/5pt]  & SGM & 26.50 & 7.62 &22.54  & 13.88 & 8.98  \\
						\cdashline{2-7}[2pt/5pt]& LinBP & 31.34 & 8.32 & 24.42& 16.40 & 8.82 \\
						\cdashline{2-7}[2pt/5pt]& Ghost & 19.8 & 6.18 &15.92 & 12.16 & 8.38  \\
						\cdashline{2-7}[2pt/5pt]& BPA & \underline{39.12} &\underline{12.74} & \underline{30.26} & \underline{22.30} & \underline{9.04}\\
						\hline
						\multicolumn{2}{c}{ BETAK (Ours)}& $\mathbf{45.48}$ & $\mathbf{14.58}$ & $\mathbf{38.18}$ & $\mathbf{26.72}$ & $\mathbf{9.68}$  \\
						\hline \multirow{5}{*}{ MI-FGSM } & N/A & 27.14 &  8.82& 23.04 & 17.52 & 9.38 \\
						\cdashline{2-7}[2pt/5pt]& SGM & 33.88& 11.30 &  31.64& 20.62 & 10.42 \\
						\cdashline{2-7}[2pt/5pt]& LinBP &40.34 & 9.92 & 31.62&21.06 & 9.52  \\
						\cdashline{2-7}[2pt/5pt]& Ghost & 30.70 & 9.62 & 24.62& 19.14 & 9.62  \\
						\cdashline{2-7}[2pt/5pt]& BPA & \underline{52.42} & \underline{17.66} & \underline{40.96}& \underline{31.34} & \underline{11.36}  \\
						\cline{1-7}
						\multicolumn{2}{c}{ BETAK (Ours)}& $\mathbf{59.80}$ & $\mathbf{22.64}$ & $\mathbf{53.44}$ & $\mathbf{36.26}$ & $\mathbf{12.52}$  \\
						\bottomrule[1.2pt] 
					\end{tabular}
				}
			\end{threeparttable}
		}
				\caption{Comparative results for ATR (\%) based on 5 defense methods with PGD or MI-FGSM as the basic attacker.} \label{tab:defense_methods}
	\end{table}

	\textbf{Evaluation with Defense Strategies.} We implement 5 representative defense methods for evaluation, including HGD~\cite{liao2018defense}, R\&P~\cite{xie2017mitigating}, NIPS-r3~\cite{kurakin2018adversarial2}, JPEG~\cite{guo2017countering} and RS~\cite{cohen2019certified}. We compare the ATR of BETAK and these methods against different defense strategies based on the PGD and MI-FGSM in Fig.~\ref{tab:defense_methods}. It can be observed that BETAK generates AEs that have consistently stronger attack performance against different defense techniques.  
	
	\begin{figure}[!t]
		\centering  
			\includegraphics[height=4.75cm,width=8.6cm,trim=6 10 10 0 , clip]{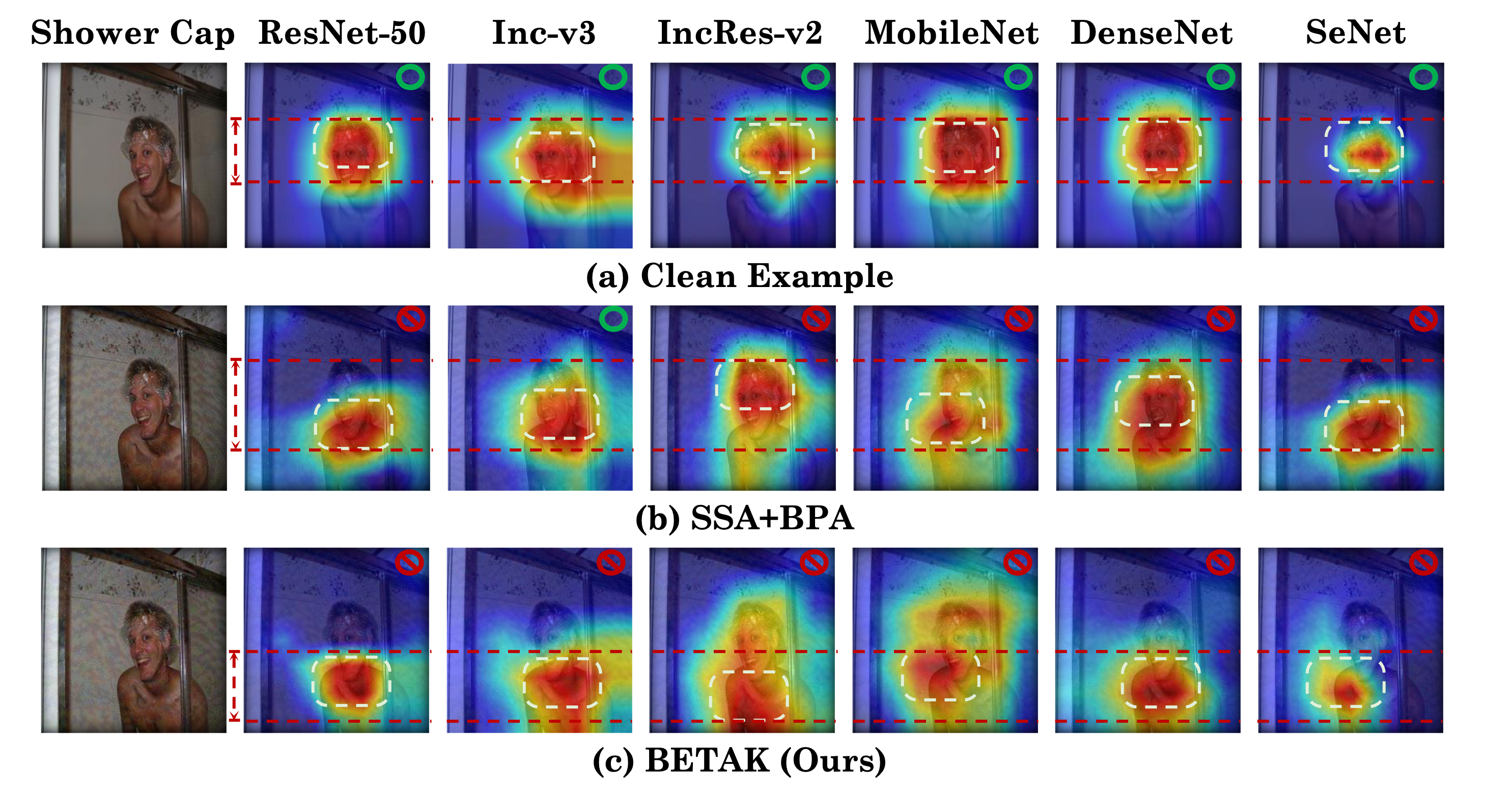}\\  
		\caption{The CAM results of the clean image (labeled as Shower Cap) in Fig.~\ref{fig:loss_landscape} and corresponding AEs generated with ResNet-50 by SSA$+$BPA and the proposed BETAK. Green and red marks represent that the image is correctly and wrongly classified, respectively.}\label{fig:fig1}%
	\end{figure}%

	\begin{table*}[htbp]
		\centering

		\scalebox{0.8}{
			\begin{threeparttable}   
				\renewcommand{\arraystretch}{1.2} 			
				\setlength{\tabcolsep}{1.1mm}{
					\begin{tabular}{ccccccccccc}
						\toprule[1.2pt] 
						Attacker & Method & Inc-v3 & IncRes-v2 & DenseNet & MobileNet & PNASNet & SENet & Inc-v$3_{ens3}$ & Inc-v$3_{ens4}$ & IncRes-v$2_{ens }$ \\
						\hline 
						PGD& BPA & $35.36 $& $30.12$ & $70.70 $& $68.90$ & $32.52$ & $42.02$ & $22.72$ & $19.28$ & $12.40$ \\
						\hline	\multirow{3}{*}{BETAK}&$+$Inc-v3$^{*}$& $\underline{50.08}_{\uparrow14.72}$ & $36.96_{\uparrow6.84}$& ${72.56}_{\uparrow1.86}$& ${73.24}_{\uparrow4.34}$ & ${36.66}_{\uparrow4.14}$ & ${45.96}_{\uparrow3.94}$& ${26.84}_{\uparrow4.12}$& ${21.78}_{\uparrow2.50}$ & ${14.08}_{\uparrow1.68}$ \\ 
						\cdashline{2-11}[2pt/5pt]&$+$IncRes-v2$^{*}$& ${46.64}_{\uparrow11.28}$ & $\mathbf{48.66}_{\uparrow18.54}$& $\underline{74.58}_{\uparrow3.88}$& $\underline{74.44}_{\uparrow5.54}$ & $\underline{41.94}_{\uparrow9.42}$ & $\underline{49.42}_{\uparrow7.40}$& $\underline{30.08}_{\uparrow7.36}$& $\underline{24.18}_{\uparrow4.90}$ & $\underline{16.82}_{\uparrow4.42}$ \\ 
						\cdashline{2-11}[2pt/5pt]&$+$Inc-v3$^{*}$, IncRes-v2$^{*}$& $\mathbf{53.34_{\uparrow17.98}}$ & $\underline{45.08}_{\uparrow14.96}$& $\mathbf{{74.68}_{\uparrow3.98}}$& $\mathbf{{74.54}_{\uparrow5.64}}$ & $\mathbf{{42.48}_{\uparrow9.96}}$ & $\mathbf{{49.94}_{\uparrow7.92}}$& $\mathbf{{30.80}_{\uparrow8.08}}$& $\mathbf{24.90_{\uparrow5.62}}$ & $\mathbf{17.60_{\uparrow5.20}}$ \\ 
						\bottomrule[1.2pt] 
					\end{tabular}
				}
			\end{threeparttable}
		}
				\caption{We report the ablation results of ATR (\%) for the proposed BETAK by adopting different combination of pseudo-victim models to design the UL subproblem. The best and second-best outcomes are designated with boldface and underline, respectively.
		} \label{tab:ablation_1}
	\end{table*}  
	
	\textbf{Visualization of Adversarial Loss Landscapes.} In Fig.~\ref{fig:loss_landscape}, we also plot the adversarial loss landscapes of different victim models for the same clean image and two adversarial images generated by SSA+BPA and BETAK in Fig.~\ref{fig:fig1}. The adversarial loss is calculated with $\mathcal{L}_{\mathtt{vic}}(\boldsymbol{u} + x\boldsymbol{\vec{\iota}}+y\boldsymbol{\vec{o}})$, where $\boldsymbol{u}$ denotes the original clean image or adversarial image, $\boldsymbol{\vec{\iota}}=\mathtt{sgn}(\nabla_{\boldsymbol{u}}\mathcal{L}_{\mathtt{atk}}(\boldsymbol{u}))$ and $\boldsymbol{\vec{o}}\sim \textrm{Rademacher}(0,0.5)$ are the sign gradient direction and random direction ($x$ and $y$ are the corresponding linear coefficients). To be general, when the original adversarial image exhibits stronger attack performance, the loss near the origin of the adversarial loss landscape should be larger and exhibit a steeper variation trend.
	
	We can easily observe that the adversarial loss landscapes of the adversarial image generated by SSA$+$BPA for different victim models are more similar to these landscapes of the clean image. In comparison, adversarial landscapes of BETAK for different victim models have greater loss and are significantly steeper than the clean image and the adversarial image of SSA$+$BPA, demonstrating stronger attack performance against the surrogate model, pseudo-victim model and two victim models. 
	
		\textbf{Analysis of the CAM Results.} To explained the mechanism of BETAK more vividly, we compare the Class Activation Mapping (CAM)~\cite{selvaraju2017grad} results of  6 victim models for the clean image and the AEs generated by SSA$+$BPA and our BETAK in Fig.~\ref{fig:fig1}. Darker red regions indicate that these pixels are more important for the classification output, and darker blue indicates that these pixel regions are less important. The white rectangular dashed box and the dashed lines highlight the regions that the model focuses on the most in both clean samples and AEs, significantly influencing the classification outputs. We can draw the following conclusions.
	\begin{itemize}
		\item Although these models have totally different network structures, they all payed much attention to similar area  of the clean image around the key item for classification.
		\item These victim models reacted absolutely different to the AE of SSA$+$BPA. In particular, Inc-v3 still correctly classified the shower cap, and both IncRes-v2 and DenseNet still payed much attention to the critical area of the AE as same as the clean image.
		\item In comparison, all the victim models misclassified the AE generated by BETAK, and they also turn their attention to totally different areas distant from the clean image. More importantly, the critical areas of the AE that these victim models focus on are concentrated in similar locations, which indicates that a better perturbation initialization is provided by BETAK.  
	\end{itemize}

	\textbf{Targeted Transfer Attack Scenario.} We evaluate the generalization performance of BETAK on the targeted transfer attacks. Practically, we follow previous works~\cite{zhao2021success} to optimize the logits loss based on the PGD and MI-FGSM attacker. In Tab.~\ref{tab:targeted_attack}, we report the ATR of BETAK against different victim models under the targeted attack scenario. It can be observed that BETAK also consistently improves the attack performance under the more challenging targeted attack scenario.
	
	\begin{figure}[!t]
		\centering 
		\begin{tabular}{c@{\extracolsep{0.1em}}}	 
			\includegraphics[width=8.5cm,height=4.8cm,trim=0 0 0 0, clip]{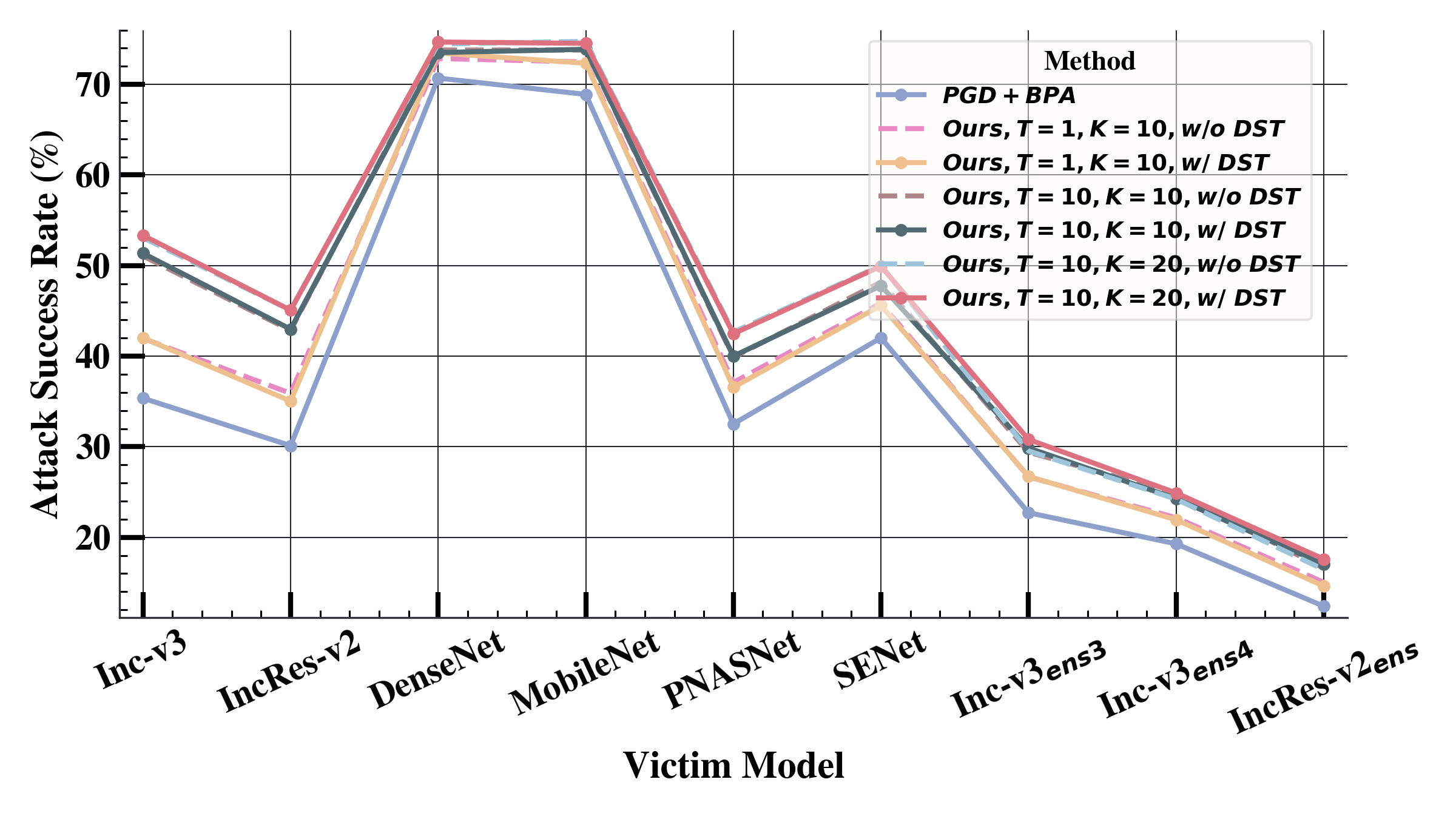}\\  
		\end{tabular}%
		\caption{Ablation results of the ATR for PGD+BPA and our BETAK framework by adjusting $T$, $K$ or removing the DST operation (denoted as w/ and w/o DST).  }\label{fig:ablation_1}%
	\end{figure}%

	\subsection{Ablation Study}
	\textbf{Influence of Pseudo-victim Models.} We conduct the ablation study to evaluate the attack performance of BETAK by selecting different pseudo-victim models to construct the UL victim attacker in Tab.~\ref{tab:ablation_1}. We have the following observations.
	\begin{itemize}
		\item We first evaluate the ATR of BETAK with a single victim model, i.e., Inception-v3 or IncRes-v2. We can easily find that even with a single pseudo-victim model participating in the UL objective, BETAK also significantly improves the ATR again the left 8 victim models.
		\item Integrating IncRes-v2 exhibits higher performance improvement compared with Inception-v3, which can be explained by the fact that IncRes-v2 is more advanced by combing ResNet and Inception-v4 with higher performance and more complex structure.
		\item When we integrate both models to construct the UL objective, BETAK obtains better attack performance across all the 9 victim models. Therefore, combining more pseudo-victim models essentially helps generate AEs with more effective feedback of the transferability, which will facilitate the generalization performance by optimizing Eq.~\eqref{eq:blo_attack}.
	\end{itemize}

	\textbf{Influence of $K$ and $T$.} We conduct the ablation study to analyze the influence of UL and LL iteration. In Fig.~\ref{fig:ablation_1}, we compare the ATR of BETAK with PGD$+$BPA by changing $T$, $K$ or removing the DST operation. As for the influence of $T$, we can observe that even 1-step update of $\boldsymbol{\delta}$ also significantly enhance the attack performance of generated AEs, which demonstrate the effectiveness of BETAK. When we adopt more UL iterations for update, the adversarial perturbation converges with a better initialization by continuously optimizing $F(\boldsymbol{\delta})$, thus the attack performance are further improved. From this perspective, implementing BETAK with larger $T$ improves the convergence results, thereby facilitating the transferability of AEs. As for the influence of $K$, the larger $K$ leads to better approximation of $\boldsymbol{\boldsymbol{\phi}}^{\star}({\boldsymbol{\delta}})$ to benefit the HGR estimation, which is demonstrated by the performance improvement of ATR with larger $K$.

\begin{figure}[!t]
		\begin{center}
			\renewcommand\arraystretch{0.8}
			\begin{tabular}{@{\extracolsep{-0.5em}}c@{\extracolsep{0.1em}}c@{\extracolsep{0.1em}}}
				\includegraphics[width=4.3cm,height=3.9cm,trim=0 10 0. 0,clip]{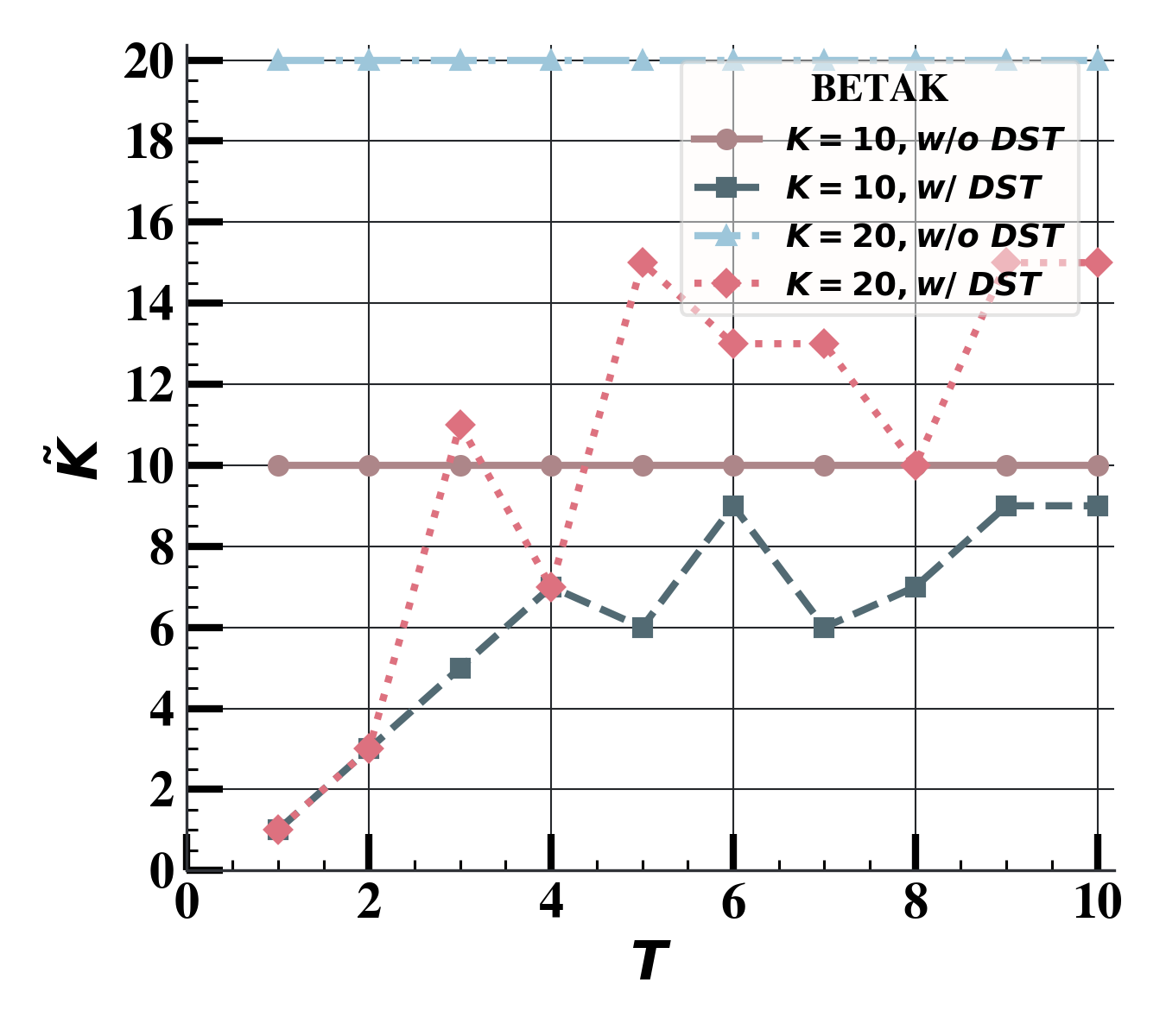}&\includegraphics[width=4.3cm,height=3.9cm,trim=0 10 0. 0,clip]{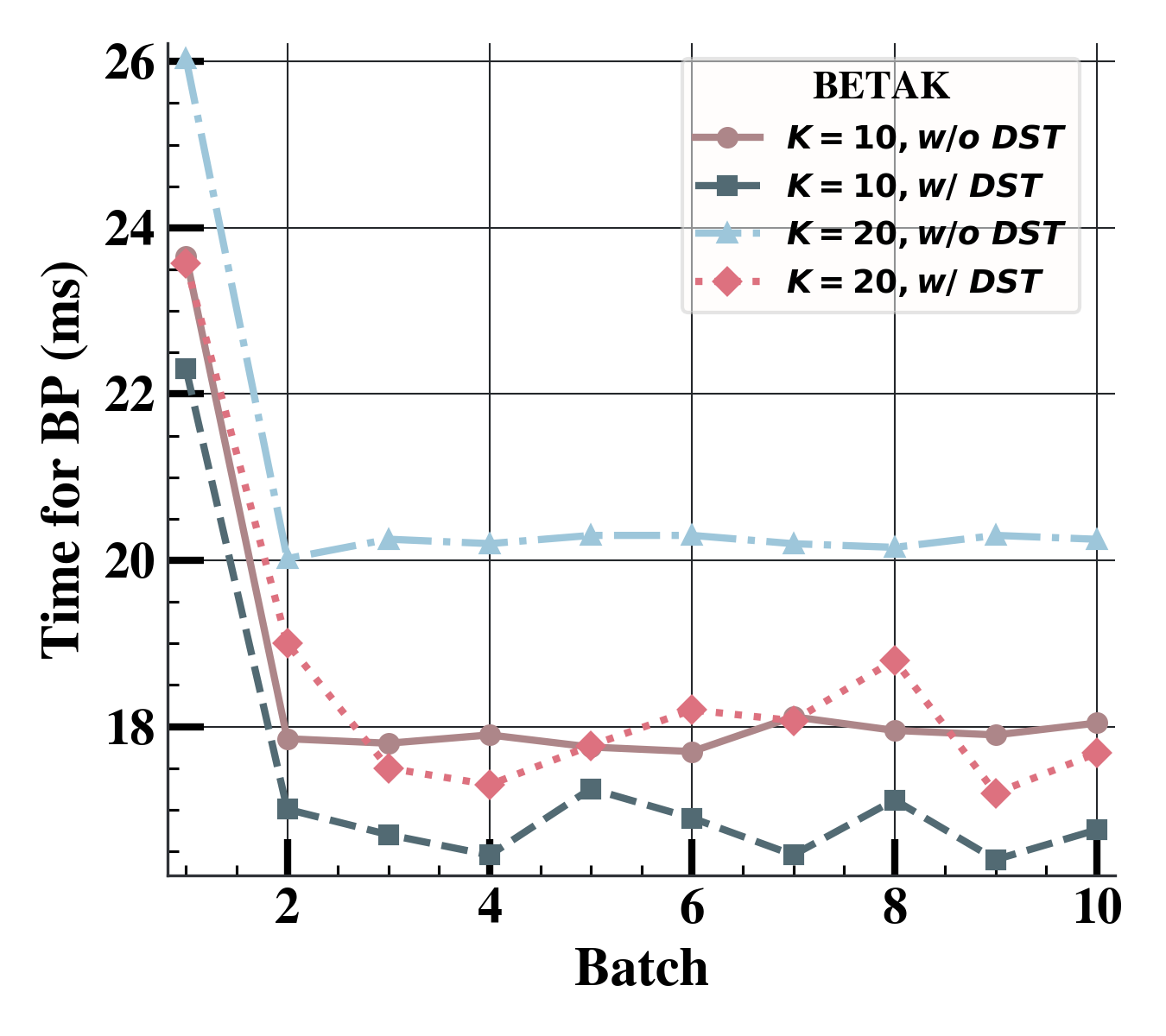}\\
			\end{tabular}
		\end{center}
		\caption{Ablation results of the time for Backpropagation (BP) for the first 10 batches of images and the average $\tilde{K}$ of DST operation as the UL iteration increases. Note that when DST operation is removed, $\tilde{K}$ is set to the same value as $K$.}\label{fig:ablation_2}
	\end{figure}

	\textbf{Influence of the DST Operation.}  From Fig.~\ref{fig:ablation_1}, it is observed that the introduction of the DST operation does not have a significant impact on the attack performance when $K=10$. However, as $K$ increases to 20, the absence of the DST operation results in the degradation of the attack performance on certain victim models compared to $K=10$. In comparison, incorporating DST operation yields the highest attack performance across various victim models. Furthermore, the influence of DST has also been analyzed with different $K$ in Tab.~\ref{fig:ablation_2}. The observation could be summarized as follows.
	
	\begin{itemize}
		\item The variation of $\tilde{K}$ with and without DST operation as $T$ increases was examined in the first subfigure. The values depicted were calculated using the average value of first 10 batches during training. It is evident that $\tilde{K}$ with DST operation are significantly reduced relative to the fixed $K$ and tend to stabilize as $T$ increases. 
		\item We also calculate the runtime time for Backpropagation (BP) across the first 10 batches. Note that higher BP time of the first batch is attributable to one-time initialization such as computation graph construction. 
		\item As indicated in the second subfigure, DST operation enables the BP cost at $K=20$ to be even lower than at $K=10$ without DST. In summary, $\tilde{K}$ with DST operation not only assists BETAK in achieving superior attack performance but also effectively reduces the computational overhead for HGR estimation.sh
	\end{itemize}

	\section{Conclusion}
	This paper introduces an initialization derived optimization paradigm to enhance the transferability of AEs, along with two efficient techniques to improve theoretical properties and computation efficiency under the BETAK framework.  Comprehensive experimental evaluations validate BETAK's effectiveness with substantial ATR increase across diverse victim models and defense techniques. 
\section*{Acknowledgments}

This work is partially supported by the National Key R\&D Program of China (No. 2022YFA1004101), the National Natural Science Foundation of China (Nos. U22B2052, 62302078 and 61936002) and the Liaoning Revitalization Talents Program (No. 2022RG04).

\bibliographystyle{named}
\bibliography{ijcai24}

\end{document}